\documentclass{Interspeech}

\interspeechcameraready

\title{
Language-Guided Contrastive Audio-Visual Masked Autoencoder\\
with Automatically Generated Audio-Visual-Text Triplets from Videos
}

\author[affiliation={1,2}]{Yuchi}{Ishikawa}
\author[affiliation={1}]{Shota}{Nakada}
\author[affiliation={1}]{Hokuto}{Munakata}
\author[affiliation={1}]{Kazuhiro}{Saito}
\author[affiliation={1}]{Tatsuya}{Komatsu}
\author[affiliation={2}]{Yoshimitsu}{Aoki}

\affiliation{}{LY Corporation}{Japan}
\affiliation{}{Keio University}{Japan}
\email{yuchi.ishikawa@lycorp.co.jp, aoki@elec.keio.ac.jp}


\usepackage{graphicx}
\usepackage{booktabs}
\usepackage{comment}
\usepackage{color}
\usepackage{multirow}
\usepackage{listings}
\usepackage{subcaption}
\usepackage{bm}
\usepackage{amsmath}

\usepackage{caption}

\begin{document}

\maketitle

\begin{abstract}
In this paper, we propose Language-Guided Contrastive Audio-Visual Masked Autoencoders (LG-CAV-MAE)
to improve audio-visual representation learning.
LG-CAV-MAE integrates a pretrained text encoder into contrastive audio-visual masked autoencoders,
enabling the model to learn across audio, visual and text modalities.
To train LG-CAV-MAE, we introduce an automatic method
to generate audio-visual-text triplets from unlabeled videos.
We first generate frame-level captions using an image captioning model and
then apply CLAP-based filtering to ensure strong alignment between audio and captions.
This approach yields high-quality audio-visual-text triplets without requiring manual annotations.
We evaluate LG-CAV-MAE on audio-visual retrieval tasks, as well as an audio-visual classification task.
Our method significantly outperforms existing approaches, achieving up to a 5.6\% improvement in recall@10 for retrieval tasks
and a 3.2\% improvement for the classification task.
\end{abstract}

\section{Introduction}

Understanding the relationship between audio and visual modalities
is crucial for various applications,
including video content analysis and autonomous systems.
Existing methods~\cite{gong2022contrastive,huang2024mavil,
georgescu2023audiovisual} have sought
to learn correspondences between auditory and visual signals
through self-supervised audio-visual learning,
achieving high performance on classification and retrieval tasks.

Multimodal learning typically requires large amounts of paired data,
which can be costly to collect.
In terms of audio-visual pairs,
videos are a relatively easy source,
as they typically contain both auditory and visual modalities.
To date, a variety of audio-visual datasets,
including AudioSet~\cite{gemmeke2017audio},
Kinetics-Sound~\cite{kay2017kinetics},
VGGSound~\cite{chen2020vggsound},
and ACAV100M\cite{lee2021acav100m},
have been released.

However, from a quality standpoint,
audio data often contains significant noise (e.g., off-screen sounds, background music),
making it challenging to collect audio-visual data with clear alignment.
Consequently, existing state-of-the-art audio-visual learning methods,
such as CAV-MAE~\cite{gong2022contrastive}, still struggle
to accurately capture fine-grained audio-visual correspondences.

To address this challenge, \cite{nakada2024deteclap} has proposed DETECLAP,
an extension of CAV-MAE.
DETECLAP is a learning framework that,
in addition to training a contrastive audio-visual masked autoencoder,
incorporates an auxiliary task of predicting object labels detected within videos.
This framework enables learning audio-visual representations
that focus on fine-grained objects rather than coarse object categories.
However, DETECLAP relies heavily on predefined object labels,
which limits its ability to learn correspondences
for objects not defined by those labels
or nuances beyond object-level features (e.g. intense vs. gentle guitar performance).

In this paper, we propose Language-Guided CAV-MAE (LG-CAV-MAE),
a simple yet effective framework for enhancing audio-visual representations.
LG-CAV-MAE extends CAV-MAE by leveraging a pretrained text encoder
to incorporate contrastive losses for audio-visual-text triplets.
This framework enables the learning of more detailed audio-visual feature representations
without relying on predefined object labels, as required by DETECLAP.

\begin{figure}[t]
    \centering
    \small
    \includegraphics[width=1.0\linewidth]{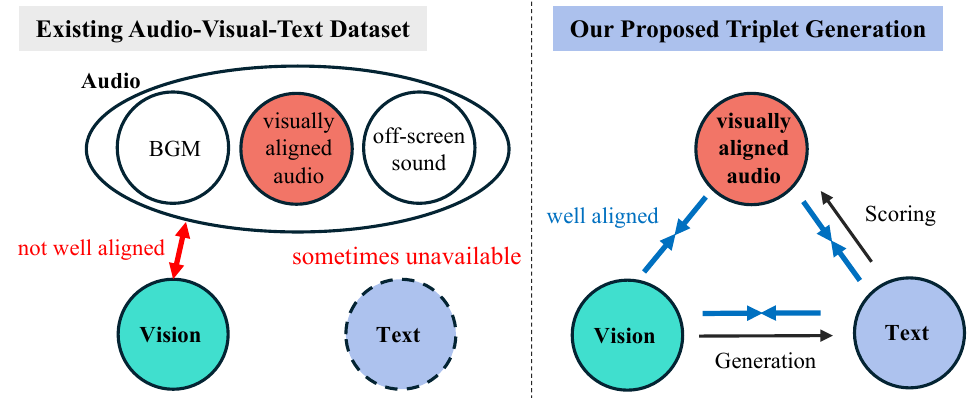}

    \caption{
    \textbf{Existing audio-visual-text data vs. our generated triplets.}
    (left) Existing datasets often lack strong alignment
    between audio and vision or have missing text.
    (right) Our proposed triplet generation method is
    based on the idea that if a caption
    generated from a video describes the audio accurately, the audio
    and visual components are also likely to correspond.
    Therefore, this method can generate well-aligned triplets.
    }
    \label{fig:top}

    \vspace{-3mm}
\end{figure}

\begin{figure*}[t]
    \centering
    \begin{subfigure}[t]{0.48\textwidth}
        \centering
        \includegraphics[width=\linewidth]{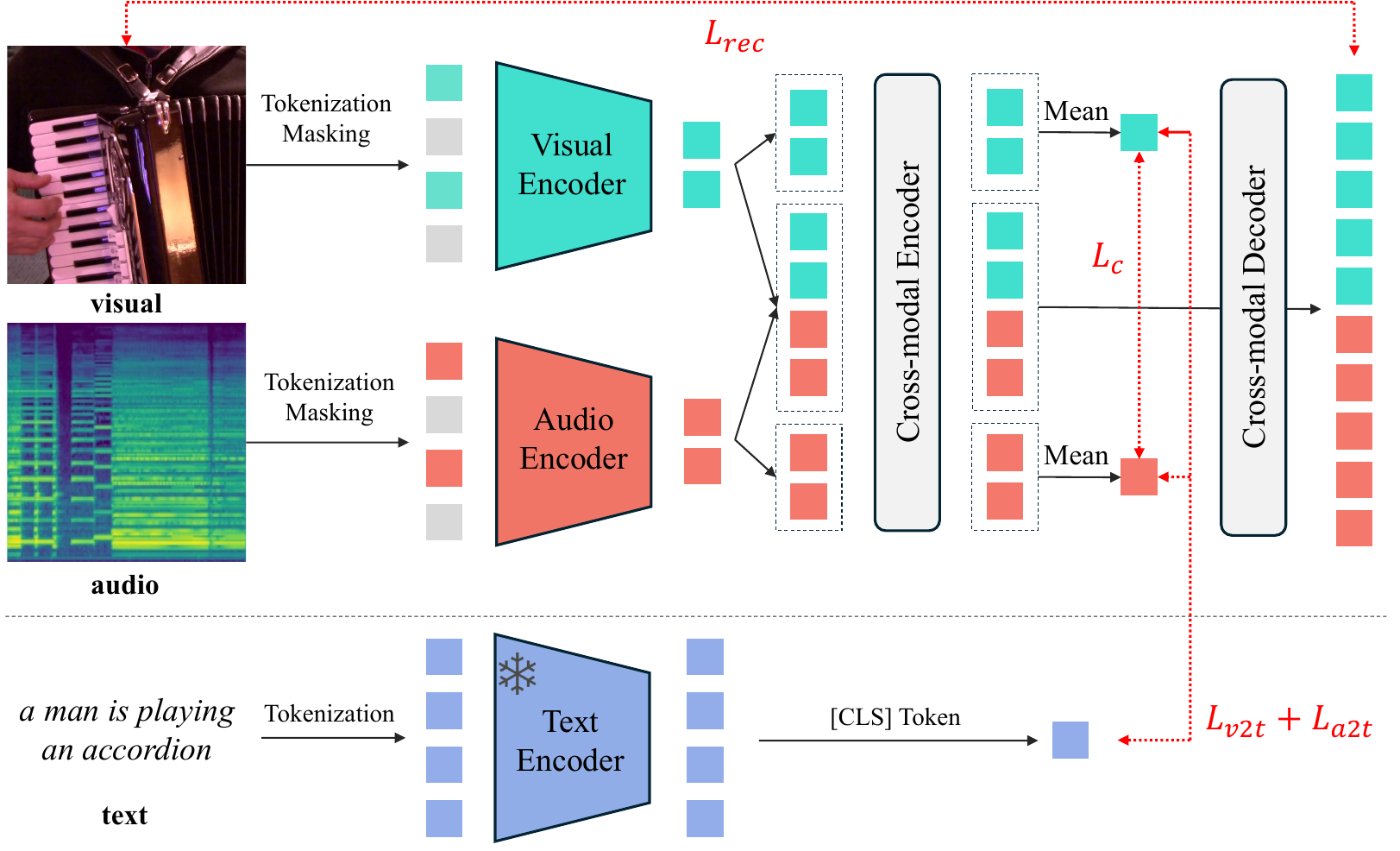}
        \caption{\textbf{Overview of our LG-CAV-MAE).}}
        \label{fig:method}
    \end{subfigure}
    \hfill
    \begin{subfigure}[t]{0.48\textwidth}
        \centering
        \includegraphics[width=\linewidth]{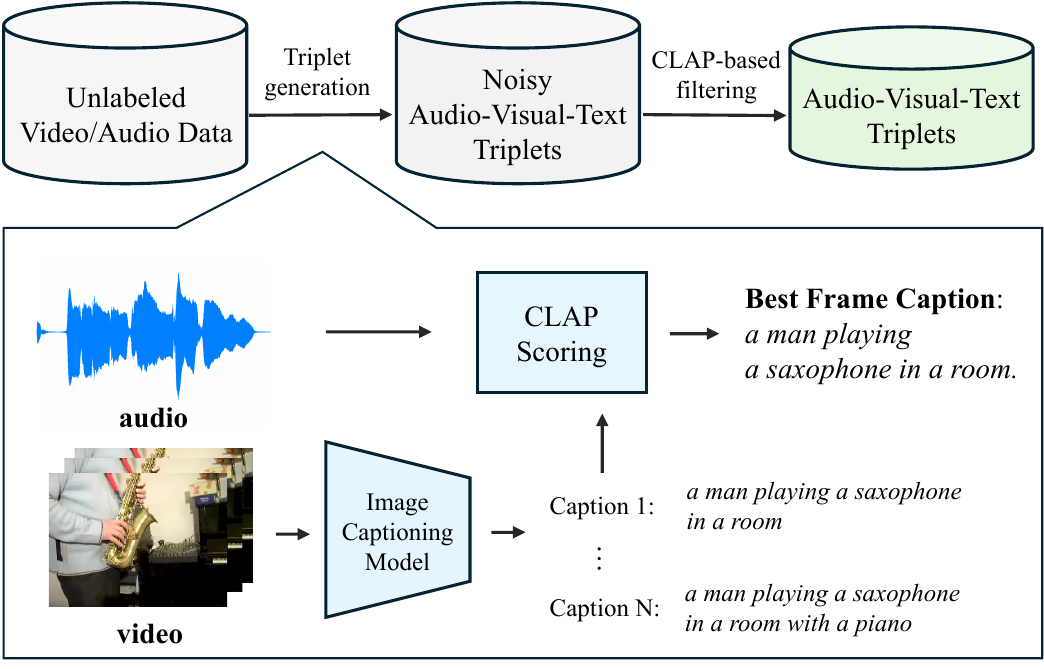}
        \caption{\textbf{Our proposed audio-visual-text triplet generation process.}}
        \label{fig:data-pipeline}
    \end{subfigure}

    \vspace{-2mm}
    \caption{
    (a) The upper part illustrates the training process of the contrastive audio-visual masked autoencoder (CAV-MAE)~\cite{gong2022contrastive}.
    The lower part shows how we introduce additional audio-text and visual-text contrastive losses
    using a pretrained text encoder, extending CAV-MAE to a tri-modal framework (LG-CAV-MAE).
    (b) For LG-CAV-MAE training, we propose to automatically generate audio-visual-text
    triplets from unlabeled videos using an image caption models and CLAP~\cite{elizalde2023clap}.
    }

    \label{fig:combined}
    \vspace{-3mm}
\end{figure*}

Although training LG-CAV-MAE requires audio-visual-text triplets,
most of existing datasets~\cite{krishna2017dense,miech2019howto100m,wang2019vatex,chen2020vggsound,lee2021acav100m}
do not provide textual descriptions nor do they offer well-aligned audio-visual pairs,
primarily due to background music and off-screen sounds (the left part of Fig.~\ref{fig:top}).
Therefore, these datasets are not suitable for our LG-CAV-MAE.

To address this problem, we further propose a method
for automatically generating captions from unlabeled videos
and filtering out those that are not descriptive on the underlying audio (the right part of Fig.~\ref{fig:top}).
This framework automatically constructs audio-visual-text triplets used for our LG-CAV-MAE training.
Our approach is based on the hypothesis that if a caption generated
from a video describes the audio accurately,
the audio and visual components are also likely to correspond.
Specifically, we first sample frames from unlabeled videos
and employ an image captioning model to generate captions for each frame.
We then compute the similarity between these generated captions
and the original video’s audio using CLAP~\cite{elizalde2023clap},
selecting instances with high similarity as audio-visual-text triplets.
This procedure allows us to construct datasets with robust audio-visual alignment
without explicitly comparing audio-visual correspondence
as in existing methods~\cite{lee2021acav100m, cheng2024avset}.
Note that, unlike in~\cite{martin2019sound,mei2024wavcaps,doh2023lp,yuan2024sound},
our method does not require metadata or tags associated with audio or videos.

In our experiments, using our automatic data generation method,
we build audio-visual-text triplets from two existing video datasets,
VGGSound~\cite{chen2020vggsound} and Kinetics700~\cite{carreira2019short}.
We then train LG-CAV-MAE with these generated triplets
and evaluate the model on audio-to-visual retrieval, visual-to-audio retrieval,
and audio-visual classification tasks.
LG-CAV-MAE achieves performance surpassing that of existing methods
without incurring any additional annotation costs.

\section{Proposed Method}

In this paper, we propose Language-Guided CAV-MAE (LG-CAV-MAE)
which is built upon CAV-MAE to enhance audio-visual representations.
We also propose to automatically generate audio-visual-text triplets from unlabeled videos,
which are subsequently used for training LG-CAV-MAE.
In the following sections, we first provide an overview of CAV-MAE,
then explain our extension of it.
Finally, we detail our method for automatically generating audio-visual-text triplets from videos.

\subsection{Preliminary: CAV-MAE}
CAV-MAE~\cite{gong2022contrastive} is a powerful self-supervised learning framework
designed to learn audio-visual representations (see the upper part of Fig.~\ref{fig:method}).
Let \(\mathcal{D} = \{(\mathbf{X}^a, \mathbf{X}^v)\}\) be a dataset consisting of paired audio and visual samples,
where $\mathbf{X}^a \in \mathbb{R}^{T_a \times F}$ is an audio log mel spectrogram
with $T_a$ time steps and $F$ frequency bins,
and $\mathbf{X}^v \in \mathbb{R}^{H \times W \times C}$ is a single video frame
of height $H$, width $W$, and $C$ channels.
CAV-MAE aims to learn audio-visual representations
by reconstructing masked regions in each modality
and aligning audio and visual representations in a shared embedding space.

In CAV-MAE training,
we first randomly mask a subset of tokens
for both the visual frame and the audio spectrogram.
For the visual input, we patchify $\mathbf{X}^v$ and
select a subset of patches indexed by $\Omega_v$.
We create a binary mask $\mathbf{M}^v$ whose entries are $0$ for masked patches
and $1$ otherwise, and obtain the masked version $\widetilde{\mathbf{X}}^v = \mathbf{M}^v \odot \mathbf{X}^v$.
Similarly, we split the audio spectrogram into patches and
choose a subset indexed by $\Omega_a$ to mask.
Let $\mathbf{M}^a$ be the corresponding binary mask;
we then have $\widetilde{\mathbf{X}}^a = \mathbf{M}^a \odot \mathbf{X}^a$.

Next, each encoder takes the corresponding masked input.
The visual encoder processes $\widetilde{\mathbf{X}}^v$
and produces a latent representation
$\mathbf{v} = \mathrm{VisualEncoder}(\widetilde{\mathbf{X}}^v) \in \mathbb{R}^{d}$
where $d$ is the dimensionality of the embedding dimension.
The audio encoder receives $\widetilde{X}^a$
and outputs $\mathbf{a} = \mathrm{AudioEncoder}(\widetilde{\mathbf{X}}^a) \in \mathbb{R}^{d}$.
Then, $\mathbf{v}$ and $\mathbf{a}$ are fed into the cross-modal encoder
to yield the following three outputs.
\begin{align}
    \label{eq:1}
    \bar{v}
    &= \rm{MeanPool}(\rm{CrossModalEncoder}(\mathbf{v}))  \in \mathbb{R}^{d} \\
    \label{eq:2}
    \bar{a}
    &= \rm{MeanPool}(\rm{CrossModalEncoder}(\mathbf{a}))  \in \mathbb{R}^{d} \\
    \label{eq:3}
    \mathbf{z}
    &= \rm{CrossModalEncoder}([\mathbf{v}, \mathbf{a}])
\end{align}

For contrastive learning, we define the contrastive loss $ \mathcal{L}_\text{c}$
using the formulation proposed in~\cite{chen2020simple}.
\begin{equation}
\label{eq:cl}
     \mathcal{L}_\text{c} = \rm{ContrastiveLoss}(\bar{a}^{+}, \bar{v}^{+}, \bar{a}^{-}, \bar{v}^{-})
\end{equation}
\noindent where $\bar{a}^{+}, \bar{v}^{+}, \bar{a}^{-}, \bar{v}^{-}$ denote 
the positive/negative audio-visual pairs in the mini-batch.

For the reconstruction task,
we pad $\mathbf{z}$ with trainable masked tokens at their original positions
to form $\mathbf{z_{pad}}$,
and feed it into the cross-modal decoder as follows: 
\begin{equation}
    [\hat{\mathbf{X}}^a, \hat{\mathbf{X}}^v] = \rm{CrossModalDecoder}(\mathbf{z_{pad}})
\end{equation}
\noindent $\mathbf{X}^a$ and $\mathbf{X}^v$ are the reconstructed audio and frame, respectively.
We then compute the reconstruction loss $ \mathcal{L}_\text{rec}$ using the mean square error as follows:
\begin{equation}
\label{eq:rec}
    \mathcal{L}_\text{rec} 
    = \frac{1}{|\Omega_v|} \sum_{i \in \Omega_v} |\mathbf{X}^v_i - \hat{\mathbf{X}}^v_i| ^2 + \frac{1}{|\Omega_a|} \sum_{j \in \Omega_a} | \mathbf{X}^a_j - \hat{\mathbf{X}}^a_j |^2
\end{equation}
\noindent The final training objective is
\begin{equation}
\label{eq:final_loss}
    \mathcal{L} = \mathcal{L}_\text{rec} + \lambda_1 \mathcal{L}_\text{c}
\end{equation}
where $\lambda_1$ balances the reconstruction and contrastive terms.

\subsection{LG-CAV-MAE}

While CAV-MAE achieves high performance in audio-visual understanding tasks,
it still struggles to capture detailed audio-visual correspondence~\cite{nakada2024deteclap}.
To learn more fine-grained audio-visual representation,
we propose a simple yet effective framework, Language-Guided CAV-MAE (LG-CAV-MAE).
LG-CAV-MAE is a tri-modal contrastive learning approach and
incorporates audio-text and visual-text contrastive losses into CAV-MAE (Fig.~\ref{fig:method}).
Let \(\mathcal{D} = \{(\mathbf{X}^a, \mathbf{X}^v, X^t)\}\) be a dataset of audio-visual-text triplets.
In addition to the CAV-MAE training process,
we feed the text $X^t$ into a pretrained text encoder and obtain the sentence-level embedding $\bar{t}$ as follows:
\begin{equation}
    \bar{t} = \rm{TextEncoder}(X^t)) \in \mathbb{R}^{d}
\end{equation}
We then apply the InfoNCE loss~\cite{oord2018representation} for the audio-text and visual-text pairs in the triplets as follows:
\begin{align}
\mathcal{L}_{\mathrm{a2t}} = \mathrm{InfoNCE}\bigl(\bar{a}^+, \bar{t}^+, \bar{a}^-, \bar{t}^-\bigr) \\
\mathcal{L}_{\mathrm{v2t}} = \mathrm{InfoNCE}\bigl(\bar{v}^+, \bar{t}^+, \bar{v}^-, \bar{t}^-\bigr)
\end{align}

\noindent The final training objective for LG-CAV-MAE is given by
\begin{equation}
\label{eq:final_loss}
    \mathcal{L} = \mathcal{L}_\text{rec} + \lambda_1 \mathcal{L}_\text{c} + \lambda_2 (\mathcal{L}_\text{a2t} + \mathcal{L}_\text{v2t})
\end{equation}
where $\lambda_2$ controls the contribution of the audio-text and visual-text contrastive losses.

\begin{table}[t]
\footnotesize

\centering
\caption{\textbf{The impact of $\lambda_2$.}
Here, LG-CAV-MAE is trained on the generated triples from VGGSound using BLIP2.
}
\vspace{-2mm}

\begin{tabular}{ccccccc}
\toprule[1.2pt]
\multirow{2}{*}{\textbf{$\lambda_2$}} & \multicolumn{3}{c}{\textbf{VGGSound}} & \multicolumn{3}{c}{\textbf{AudioSet20K}} \\
 & \textbf{R@1} & \textbf{R@5} & \textbf{R@10} & \textbf{R@1} & \textbf{R@5} & \textbf{R@10} \\
\midrule[0.5pt]
\multicolumn{7}{l}{\textbf{\textit{Audio-to-Visual Retrieval}}} \\
0.001                    & 13.9 & 36.8 & 47.3 & 7.6 & 19.1 & 26.0 \\
0.005                    & 14.6 & 37.2 & 48.3 & 7.4 & 18.7 & 25.4 \\
0.01                     & \textbf{17.7} & \textbf{41.2} & \textbf{52.2} & \textbf{8.3} & \textbf{20.8} & \textbf{29.8} \\
0.05                     & 14.4 & 36.6 & 47.3 & 6.9 & 18.7 & 25.9 \\
0.1                      & 12.0 & 35.1 & 45.9 & 6.5 & 17.7 & 24.2 \\
\midrule[0.5pt]
\multicolumn{7}{l}{\textbf{\textit{Visual-to-Audio Retrieval}}} \\
0.001                    & 16.2 & 41.8 & 51.8 & 7.4 & 20.5 & 28.3 \\
0.005                    & 14.6 & 37.2 & 48.3 & 7.7 & 20.0 & 28.1 \\
0.01                     & \textbf{18.1} & \textbf{43.9} & \textbf{54.5} & \textbf{8.3} & \textbf{20.8} & \textbf{29.8} \\
0.05                     & 16.6 & 40.5 & 50.5 & 6.9 & 18.7 & 25.9 \\
0.1                      & 14.5 & 38.4 & 48.9 & 6.5 & 17.7 & 24.2 \\
\bottomrule[1.2pt]
\end{tabular}
\label{tab:lambda2}
\vspace{-2mm}
\end{table}

\subsection{Automatic Audio-Visual-Text Triplet Generation}

Training LG-CAV-MAE requires high-quality audio-visual-text triplets.
Although there are many datasets containing audio-visual-text 
triplets~\cite{krishna2017dense,miech2019howto100m,wang2019vatex},
most of them do not offer well-aligned audio-visual pairs.
To address this problem,
we propose an automatic method for constructing audio-visual-text triplets
from unlabeled videos (as shown in Fig.~\ref{fig:data-pipeline}).
This method ensures high-quality alignment among the three modalities
while eliminating the need for manual annotations.
The triplet generation process consists of the following steps:

\noindent \textbf{Video Frame Captioning:}
Given an unlabeled video that contains an audio track,
we extract frames and apply an image captioning model to generate captions.
This produces multiple candidate captions corresponding to different frames.
In our experiments, we generate captions for frames per second.
As an image encoder, we validate BLIP2~\cite{li2023blip} and LLaVa1.5~\cite{liu2024improved}.

\noindent \textbf{Audio-Visual Alignment Scoring and Caption Selection:}
We pair the original video audio with each generated frame caption.
To assess the alignment between audio and text,
we employ CLAP~\cite{elizalde2023clap} to compute the semantic similarity between the audio and each caption.
Among the generated captions, the one with the highest CLAP score is selected
as the most representative textual description of the audio-visual content.
We then combine the selected caption with the corresponding video and audio
to form an audio-visual-text triplet.

\noindent \textbf{Filtering Triplets:}
Although this process can yield audio-visual-text triplets, 
there may still be noisy or weakly aligned samples. 
To remove them, we compute the CLAP scores for all generated triplets 
and select the top $k\%$ of samples with the highest scores, 
where $k$ is chosen from $\{50, 30, 10\}$ in our experiments.
Through empirical evaluation, we compare these three thresholds 
to determine which offers the best balance for training LG-CAV-MAE.

\section{Experiments}

\begin{table}[t]
\footnotesize

\centering
\caption{
\textbf{Effects of caption models on our triplet generation.}
Here, we train LG-CAV-MAE on VGGSound.
}
\vspace{-2mm}

\begin{tabular}{lcccccc}
\toprule[1.2pt]
\multirow{2}{*}{\textbf{Model}} & \multicolumn{3}{c}{\textbf{VGGSound}} & \multicolumn{3}{c}{\textbf{AudioSet20K}} \\
 & \textbf{R@1} & \textbf{R@5} & \textbf{R@10} & \textbf{R@1} & \textbf{R@5} & \textbf{R@10} \\
\midrule[0.5pt]
\multicolumn{7}{l}{\textbf{\textit{Audio-to-Visual Retrieval}}} \\
BLIP2                               & \textbf{17.7} & \textbf{41.2} & \textbf{52.2} & 8.3 & \textbf{20.8} & \textbf{29.8} \\
LLaVa1.5                            & 17.3 & 39.3 & 50.9 & \textbf{8.6} & 20.6 & 28.9 \\
\midrule[0.5pt]
\multicolumn{7}{l}{\textbf{\textit{Visual-to-Audio Retrieval}}} \\
BLIP2                               & 18.1 & \textbf{43.9} & \textbf{54.5} & 8.3 & 20.8 & 29.8 \\
LLaVa1.5                            & \textbf{18.3} & 42.1 & 52.6 & \textbf{8.7} & \textbf{22.3} & \textbf{30.3} \\
\bottomrule[1.2pt]

\end{tabular}
\label{tab:caption_model}

\vspace{-2mm}
\end{table}

\subsection{Experimental settings}

To evaluate the efficacy of our LG-CAV-MAE and the automatic generation of audio-visual-text triplets,
we pretrain LG-CAV-MAE with the generated audio-visual-text triplets and
assess its performance on audio-to-visual and visual-to-audio retrieval tasks,
as well as on audio-visual classification.

\noindent \textbf{Dataset: }
For data generation and pretraining,
we use VGGSound~\cite{chen2020vggsound} and Kinetics700~\cite{carreira2019short}.
VGGSound is a large-scale audio-visual dataset containing 200k videos.
Kinetics700 consists of 650k videos spanning 700 human action classes.
Although VGGSound provides relatively well-aligned audio-visual pairs,
Kinetics700 contains some audio noise, leading to lower-quality audio-visual pairs.
We apply our data generation framework to Kinetics700
and show how effectively our framework filter out such noisy samples.
Note that we do not employ AudioSet~\cite{gemmeke2017audio} for pre-training,
as was done in~\cite{gong2022contrastive},
because we do not have access to the full AudioSet dataset.

\noindent \textbf{Evaluation Metrics: }
For evaluation on audio-to-visual and visual-to-audio retrieval tasks,
we use AudioSet20k~\cite{gemmeke2017audio}, along with VGGSound.
Following previous works~\cite{gong2022contrastive,nakada2024deteclap},
we report recall@K (R@K) for $K = \{1, 5, 10\}$,
which measures the proportion of correctly retrieved items within the top $K$ results.

For the audio-visual classification task,
we use AudioSet20k, which includes 517 category labels,
and VGGSound, which contains 309 category labels.
While AudioSet20k is designed for multi-label classification,
VGGSound serves as a benchmark for multi-class classification.
In this experiment, we finetune the pretrained model using the labels
and report mean average precision (mAP) for AudioSet20k and classification accuracy for VGGSound.

\begin{table}[t]
\footnotesize
\scriptsize

\centering
\caption{
\textbf{Perfromance of LG-CAV-MAE trained on 
the generated triplets from VGGSound (VS) and Kinetics700 (K).}
The numbers in parentheses show the proportion of
triplets selected through CLAP filtering.
}
\vspace{-2mm}

\begin{tabular}{lcccccc}
\toprule[1.2pt]
\multirow{2}{*}{\textbf{Dataset}} & \multicolumn{3}{c}{\textbf{VGGSound}} & \multicolumn{3}{c}{\textbf{AudioSet20K}} \\
 & \textbf{R@1} & \textbf{R@5} & \textbf{R@10} & \textbf{R@1} & \textbf{R@5} & \textbf{R@10} \\
\midrule[0.5pt]
\multicolumn{7}{l}{\textbf{\textit{Audio-to-Visual Retrieval}}} \\
K (10\%)                    & 4.4  & 11.8 & 17.4 & 4.4 & 11.8 & 17.4 \\
K (30\%)                    & 5.4  & 16.4 & 21.3 & 3.5 & 9.9  & 15.1 \\
K (50\%)                    & 6.0  & 16.6 & 23.5 & 4.1 & 11.1 & 16.9 \\
K                           & 8.2  & 21.0 & 28.0 & 8.0 & 17.1 & 23.2 \\
VS                          & 17.7 & 41.2 & 52.2 & 8.3 & 20.8 & 29.8 \\
VS + K (10\%)               & 18.3 & 41.5 & 52.8 & 9.9 & 24.0 & 31.4 \\
VS + K (30\%)               & \textbf{18.9} & \textbf{43.3} & \textbf{54.5} & 10.6& \textbf{25.9} & \textbf{35.4} \\
VS + K (50\%)               & 18.0 & 41.7 & 53.0 & \textbf{10.8}& 24.5 & 33.1 \\
\midrule[0.5pt]
\multicolumn{7}{l}{\textbf{\textit{Visual-to-Audio Retrieval}}} \\
K (10\%)                    & 3.1  & 10.3 & 15.2 & 3.1  & 10.3& 15.2 \\
K (30\%)                    & 4.9  & 14.2 & 19.8 & 2.4  & 8.6 & 11.9 \\
K (50\%)                    & 5.7  & 15.1 & 21.5 & 3.8  & 9.8 & 14.2 \\
K                           & 7.4  & 19.1 & 25.5 & 6.1  & 13.7& 19.0 \\
VS                          & 18.1 & 43.9 & 54.5 & 8.3 & 20.8 & 29.8 \\
VS + K (10\%)               & 20.6 & 44.6 & 56.2 & 9.8 & 23.8 & 32.8 \\
VS + K (30\%)               & \textbf{20.8} & \textbf{46.3} & \textbf{56.5} & \textbf{11.3} & \textbf{26.0} & \textbf{34.0} \\
VS + K (50\%)               & 18.8 & 45.0 & 55.0 & 11.0& 25.7 & 33.6 \\
\bottomrule[1.2pt]

\end{tabular}
\label{tab:dataset}
\vspace{-1mm}
\end{table}

\begin{table}[t]
\scriptsize

\centering
\caption{
\textbf{Effectiveness of CLAP filtering in our triplet generation.}
Here, VS = VGGSound, K = Kinetics700, and "rand. 30\%" indicates
that 30\% of the data was randomly selected.
}
\vspace{-2mm}

\begin{tabular}{lcccccc}
\toprule[1.2pt]
\multirow{2}{*}{\textbf{Dataset}} & \multicolumn{3}{c}{\textbf{VGGSound}} & \multicolumn{3}{c}{\textbf{AudioSet20K}} \\
 & \textbf{R@1} & \textbf{R@5} & \textbf{R@10} & \textbf{R@1} & \textbf{R@5} & \textbf{R@10} \\
\midrule[0.5pt]
\multicolumn{7}{l}{\textbf{\textit{Audio-to-Visual Retrieval}}} \\
VS                          & 17.7 & 41.2 & 52.2 & 8.3 & 20.8 & 29.8 \\
+ K (30\%)                  & \textbf{18.9} & \textbf{43.3} & \textbf{54.5} & \textbf{10.6}& \textbf{25.9} & \textbf{35.4} \\
+ K (rand. 30\%)            & 17.9 & 41.6 & 53.6 & 10.0 & 24.1 & 31.7 \\
\midrule[0.5pt]
\multicolumn{7}{l}{\textbf{\textit{Visual-to-Audio Retrieval}}} \\
VS                          & 18.1 & 43.9 & 54.5 & 8.3 & 20.8 & 29.8 \\
+ K (30\%)                  & \textbf{20.8} & \textbf{46.3} & \textbf{56.5} & \textbf{11.3} & \textbf{26.0} & \textbf{34.0} \\
+ K (rand. 30\%)            & 19.7 & 44.9 & 55.8 & 9.9 & 25.8 & 32.8 \\
\bottomrule[1.2pt]

\end{tabular}
\label{tab:filtering}
\vspace{-2mm}
\end{table}

\begin{table}[t]
\footnotesize
\scriptsize

\centering
\caption{
\textbf{Comparison with existing methods on audio-to-visual and visual-to-audio retrieval tasks.}
All models are trained on VGGSound except for Ours (+K 30\%),
which is trained on VGGSound plus the top 30\% of Kinetics700 triplets selected by CLAP scores.
Note that, since the original CAV-MAE~\cite{gong2022contrastive} was trained on AudioSet, our results are not directly comparable to those reported in the original paper.
}
\vspace{-2mm}

\begin{tabular}{lcccccc}
\toprule[1.2pt]
\multirow{2}{*}{\textbf{Method}} & \multicolumn{3}{c}{\textbf{VGGSound}} & \multicolumn{3}{c}{\textbf{AudioSet20K}} \\
 & \textbf{R@1} & \textbf{R@5} & \textbf{R@10} & \textbf{R@1} & \textbf{R@5} & \textbf{R@10} \\
\midrule[0.5pt]
\multicolumn{7}{l}{\textbf{\textit{Audio-to-Visual Retrieval}}} \\
CAV-MAE~\cite{gong2022contrastive} & 15.1 & 36.6 & 48.0 & 7.4 & 18.4 & 25.7 \\
DETECLAP~\cite{nakada2024deteclap} & 15.2 & 39.2 & 49.5 & 7.9 & 20.4 & 27.7 \\
Ours                               & 17.7 & 41.2 & 52.2 & 8.3 & 20.8 & 29.8 \\
Ours (+K 30\%)                     & \textbf{18.9} & \textbf{43.3} & \textbf{54.5} & \textbf{10.6} & \textbf{25.9} & \textbf{35.4} \\
\midrule[0.5pt]
\multicolumn{7}{l}{\textbf{\textit{Visual-to-Audio Retrieval}}} \\
CAV-MAE~\cite{gong2022contrastive} & 15.7 & 39.6 & 50.5 & 7.4 & 20.0 & 28.9 \\
DETECLAP~\cite{nakada2024deteclap} & 17.4 & 43.0 & 51.7 & 8.2 & 22.1 & 29.7 \\
Ours                               & 18.1 & 43.9 & 54.5 & 8.3 & 20.8 & 29.8 \\
Ours (+K 30\%)                     & \textbf{20.8} & \textbf{46.3} & \textbf{56.5} & \textbf{11.3}& \textbf{26.0} & \textbf{34.0} \\
\bottomrule[1.2pt]

\end{tabular}
\label{tab:audio-visual-retrieval}

\vspace{-1mm}
\end{table}

\begin{table}[ht]
\small
\footnotesize

\centering
\caption{\textbf{Audio-visual classification performance comparison on the VGGSound and AudioSet20K datasets.}}

\vspace{-2mm}

\begin{tabular}{lcc}
\toprule[1.2pt]
\multirow{2}{*}{\textbf{Method}} & \textbf{VGGSound} & \textbf{AudioSet20K} \\
 & \textbf{Accuracy} & \textbf{mAP} \\
\midrule[0.5pt]
AudioSlowFast \cite{xiao2020audiovisual} & 52.5 & - \\
Audio-MAE \cite{huang2022masked}                      & -    & 37.0 \\
CAV-MAE~\cite{gong2022contrastive}       & 58.9 & 38.4 \\
DETECLAP~\cite{nakada2024deteclap}       & 59.5 & 39.6 \\
\midrule[0.5pt]
LG-CAV-MAE (VS)                          & 59.2 & 41.1 \\
LG-CAV-MAE (VS + K 30\%)                 & \textbf{60.1} & \textbf{42.8} \\
\bottomrule[1.2pt]
\end{tabular}
\label{tab:audio-visual-classification}

\vspace{-2mm}
\end{table}

\noindent \textbf{Implementation details. }
Our experimental setting is mostly aligned with that of~\cite{nakada2024deteclap}.
We set the frame size $H \times W$ as $224 \times 224$,
the time steps of the audio spectrogram $T_a$ as 1024,
and the frequency bins $F$ as 128.
For training, we use Adam~\cite{DBLP:journals/corr/KingmaB14} as the optimizer, with a learning rate of 0.0005.
As the pretrained text encoder,
we use the 2023 version of CLAP~\cite{elizalde2023clap}.
Since the embedding dimension differs between CLAP and LG-CAV-MAE,
we feed the audio and visual embeddings into a two-layer MLP to match the dimensionality.

\subsection{Results}

\noindent \textbf{Ablation Studies:}
Tab.~\ref{tab:lambda2} investigates the impact of \(\lambda_2\).
Similarly to $\lambda_1$ in the CAV-MAE loss,
we found that a relatively small contrastive loss weight of 0.01 is preferable.
Hence, we fix $\lambda_2 = 0.01$ for subsequent experiments.

Tab.~\ref{tab:caption_model} compares performance
when using different captioning models in our triplet generation framework.
Although LLaVa1.5 is a better captioning model than BLIP2,
it did not improve performance in our approach.
Therefore, we use capition generated by BLIP2 in all subsequent experiments.

Tab.~\ref{tab:dataset} compares the performance of LG-CAV-MAE
trained on combined triplets from two different datasets.
When the model is trained on both VGGSound and Kinetics,
its performance is superior to that trained solely on Kinetics700 or VGGSound.
The results also indicate that the amount of data greatly influences performance when using Kinetics700.
In contrast, when the model is trained on both Kinetics700 and VGGSound,
simply using more Kinetics700 samples does not necessarily yield better results.
By combining the top 30\% of Kinetics700 triplets with VGGSound,
we achieve an improvement of more than 2\% compared to that trained with VGGSound alone.
From this point forward, when using Kinetics700 for our framework,
we choose the top 30\% of triplets.

Table~\ref{tab:filtering} highlights how effective CLAP filtering
is compared to random sampling.
As shown, combining VGGSound with randomly sampled data yields smaller performance gains than
when using CLAP filtering.
This suggests that CLAP filtering effectively removes noisy audio-text pairs,
thereby improving the overall quality of audio-visual-text triplets.

\noindent \textbf{Comparison with Existing Methods: }
Tab.~\ref{tab:audio-visual-retrieval} presents the performance comparison
for audio-to-visual and visual-to-audio retrieval tasks.
Our LG-CAV-MAE achieves improvements of +2.7\% and +2.8\% in R@10,
respectively, over existing methods.
Furthermore, by incorporating Kinetics700,
we gain up to an additional 5.6\% and 5\% improvement in audio-to-visual and visual-to-audio retrieval,
compared to the existing methods.

Tab.~\ref{tab:audio-visual-classification} compares the performance for audio-visual classification.
Though the performance gain on VGGSound is limited,
our framework achieves +3.2\% improvement on AudioSet20K.

\section{Conclusion}

In this paper, we proposed LG-CAV-MAE, a tri-modal extension of CAV-MAE
that incorporates audio-text and visual-text contrastive losses. 
To address the limited availability of high-quality audio-visual-text data, 
we introduced an automatic method for generating audio-visual-text triplets 
from unlabeled videos via image captioning and CLAP alignment scoring. 
Our experiments demonstrate that filtering these triplets using CLAP
effectively removes noisy audio-text pairs,
with the top 30\% samples from Kinetics700
providing significant performance gains when combined with VGGSound.
As a result, LG-CAV-MAE outperforms existing audio-visual learning approaches,
showing substantial improvements on both audio-visual retrieval tasks and audio-visual classification.

\newpage

\bibliographystyle{IEEEtran}
\bibliography{main}

\end{document}